\documentclass[12pt]{article}
\usepackage[pdftex]{graphicx,graphics}
\usepackage{grffile}
\usepackage{subfigure}
\usepackage{setspace}
\usepackage[left=1in,right=1in,top=1in,bottom=1in]{geometry}
\usepackage{times}
\usepackage{natbib}
\usepackage{epstopdf}
\usepackage[font={scriptsize}]{caption}
\usepackage{bbm}
\usepackage{wrapfig}
\usepackage{amsmath, amsthm, amssymb}

\makeatletter
\renewcommand\section{\@startsection{section}{1}{\z@}%
                                  {-3.5ex \@plus -1ex \@minus -.2ex}%
                                  {2.3ex \@plus.2ex}%
                                  {\normalfont\bfseries}}
\renewcommand\subsection{\@startsection{subsection}{1}{\z@}%
                                	{-3.5ex \@plus -1ex \@minus -.2ex}%
                                	{2.3ex \@plus.2ex}%
                                	{\normalfont\bfseries}}
\renewcommand\subsubsection{\@startsection{subsubsection}{1}{\z@}%
                                	{-3.5ex \@plus -1ex \@minus -.2ex}%
                                	{2.3ex \@plus.2ex}%
                                	{\normalfont\bfseries}}
\newcommand{\argmin}{\operatornamewithlimits{argmin}}
                                
\makeatother
\begin{document}
\noindent
\footnotesize{
\noindent
\textit{Proceedings of the 2015 INFORMS Workshop on Data Mining and Analytics\\
\noindent
M.~G.~Baydogan, S.~Huang, A.~Oztekin, eds.}}

\vspace{0.1in}
\begin{center}
    {\large\bf Learning to shortcut and shortlist order fulfillment deciding}\\
    \vspace{0.3in}
\textbf{Brian Quanz, Ajay Deshpande, Dahai Xing and Xuan Liu}

IBM Research, IBM T. J. Watson Research Center, Yorktown Heights, NY\\
\texttt{\{blquanz,ajayd,dxing,xuanliu\}@us.ibm.com}\\
\vspace{0.2in}
\end{center}

\begin{center}
    {\bf Abstract}\\
\end{center}
\noindent
With the increase of order fulfillment options and business objectives taken into consideration in the deciding process, order fulfillment deciding is becoming more and more complex.  For example, with the advent of ship from store retailers now have many more fulfillment nodes to consider, and it is now common to take into account many and varied business goals in making fulfillment decisions.  With increasing complexity, efficiency of the deciding process can become a real concern.  Finding the optimal fulfillment assignments among all possible ones may be too costly to do for every order especially during peak times.
In this work, we explore the possibility of exploiting regularity in the fulfillment decision process to reduce the burden on the deciding system.  By using data mining we aim to find patterns in past fulfillment decisions that can be used to efficiently predict most likely assignments for future decisions.  Essentially, those assignments that can be predicted with high confidence can be used to shortcut, or bypass, the expensive deciding process, or else a set of most likely assignments can be used for shortlisting - sending a much smaller set of candidates for consideration by the fulfillment deciding system.

\noindent {\bf Keywords:}
 order fulfillment, data mining, predictive optimization, machine learning

\section{Introduction}
In recent years, omni-channel retailers are utilizing larger quantities and varieties of nodes in their supply networks for fulfillment of e-commerce orders with the goal of reducing fulfillment cost, in particular their large store networks.  This raises new challenges as there are many more such nodes e.g., hundreds vs. 3-4, and further these were not designed for fulfilling e-commerce orders.  As a result many more factors and objectives must also be taken into account in addition to just total shipping cost, such as inventory availability, clearance related savings, node performance, etc.
With the increase of order fulfillment options and business objectives taken into consideration in the deciding process, order fulfillment deciding - i.e., deciding which nodes to assign which parts of an order to - is becoming more and more complex.  With increasing complexity, efficiency of the deciding process becomes a real concern.  Finding the optimal fulfillment assignments among all possible ones may be too costly to do for every order, especially during peak seasons and times.  Furthermore  retailers do not want to incur the cost for increased resources to adequately handle the peak load as it is only needed during the short peak seasons.  With the much larger fulfillment networks enabled, retailers' existing fulfillment deciding engines may not be able to adequately handle the increased load during peak season times and are forced to either revert to sub-optimal fulfillment assignments that result in higher fulfillment cost, or delay and cancel orders, severely damaging customer relationships. 

In this work we propose a new approach by which the load can be reduced on the fulfillment deciding engine, essentially by shortcutting the fulfillment deciding process to avoid the complex, time-consuming full optimization problem of determining the best of all possible assignments.  We introduce a novel shortcutting component in the fulfillment deciding process that analyzes an order before it is passed to the deciding engine. The idea is that by utilizing and learning from the past order fulfillment decisions we can find cases where the fulfillment decisions for future orders will likely follow similar patterns, and thereby greatly simplify the deciding process in those cases – e.g., by having to consider only a much smaller set of possible assignments.  Cases where the full fulfillment deciding process is likely unnecessary are accurately identified and routed to bypass the heavy-weight deciding process.  With this approach, we aim to enable retailers to handle both peak and non-peak season loads with less resources, and also reduce unnecessary utilization of the resources they do have, freeing them to be used for other tasks.  Additionally companies can also see what patterns there are in their order fulfillment, better understand how their orders are getting fulfilled, and make other supply chain changes and decisions based on this as well.

In general, various types of shortcutting are possible.  There are three shortcutting use cases in particular that motivate this work.
The first is determining / predicting when an order will be split.  Examination of a sample of peak season data from one retailer revealed that the majority of orders were not split in the end by the fulfillment deciding engine - around 75\%.  Identifying even a portion of these orders fairly accurately would allow significantly reducing the burden on the fulfillment deciding engine, since if we can say in advance an order doesn't need to be split we can simply check the total cost of shipping all order lines at each node – massively reducing the number of assignment options under consideration and greatly simplifying the solving process (e.g., linear scan of the nodes).  The second use case is repeated identical orders for ``hot'' items.  If an order arrives that is very similar to one that was recently decided, we can in many cases safely assign the order directly to the same nodes with minimal checking, thus bypassing the full-blown deciding process. This case arises quite commonly especially during peak seasons where hot or sales items are ordered at a high frequency.  The third use case is short-listing - i.e. in some cases it may be possible to predict with high probability a reduced set of nodes for which the fulfillment engine will end up assigning the order lines among, even for split orders.  In this case we may find patterns of key features that let us greatly reduce the set of nodes under consideration.

In this work we focus on the first use case, predicting order splitting.  We pick order splitting to explore first because we suspected some benefit is possible.  For example, if there is a single orderline, even with multiple items, it is probably less likely to get split.  If the total quantity in the order is lower we suspected there would be less chance of split.  If an estimate of the minimum total cost for not splitting is relatively low, we could also suspect an order won't be split.  In other words, we have suspected cases where it seems likely we could predict most of the time whether the order would be split or not, and this is where data mining / machine learning comes into play.  We use this to quantify such cases and tell us when these cases occur and exactly how likely the split will happen given the different factors.  Additionally data mining helps us to understand what are the key factors related to the order split outcomes.

We evaluate our approach on a set of real order and fulfillment network data during a peak season from a major retailer, and a prototype multi-objective fulfillment optimizer that performs the fulfillment deciding.  With the features we came up with and regularized predictive models fitted to the data, we found that we can predict order splitting by the fulfillment deciding engine with high accuracy, over 94\% on average.  Furthermore if we are willing to restrict the prediction to only the most confidently predicted cases, we can achieve on average over 99\% accuracy while still classifying almost 60\% of multi-item orders.  Additionally we explore the relationships between the features and order splitting, and what the key factors were in the predictive models.  We see this work as a promising proof of concept and a first step at validating the general idea of optimizer short-cutting, and hope to continue to extend the approach and shortcutting use cases addressed moving forward.
\section{Literature review}
In this section, we review the literature in the following three areas: e-commerce order fulfillment optimization, shipment split prediction, and data driven optimization. We by no means provide a comprehensive review in each area within this section, but only a discussion of the papers related to our research.

Order sourcing and fulfillment optimization is critical to e-commerce operations. It touches every aspect of business operations, such as inventory management, warehouse operations, and transportation management etc. Many industry solutions have been developed, and there are a few academic work in the literature. \cite{Chan:2006} focus on the optimization of order fulfillment reliability in a multi-echelon distribution network. They propose an approach integrating genetic algorithms with an analytic hierarchy process to enable multi-criterion optimization, and probabilistic analysis to capture uncertainties. \cite{Xu:2008} examine the potential benefits from periodically re-evaluating the real-time order-assignment decisions. They construct near-optimal heuristics for the re-assignment for a large set of customer orders with the objective to minimize the total number of shipments. \cite{Acimovic:2014} proposes implementable, i.e., computationally tractable and relatively intuitive, solutions for both order fulfillment and inventory replenishment problems. All of the previous work relies on the traditional operations research approach to solve the order fulfillment problem,  there is a lack of literature looking from the machine learning angle and leveraging the rich data sources to explore more innovative approaches.

Minimizing shipment cost is the most important objective in order fulfillment optimization. \cite{Xu:2008} analyze the real world data from a major e-retailer and show that the order split rate and the number of shipments are very good proxies to estimate shipment cost.  A cluster of literature addresses shipment volume split.  \cite{Surti:2005} uses linear regression model to predict freight volume splits among different transportation modes such as ground, air and water. \cite{Hwanga:2014} develop a binomial logit market share model to predict interregional freight modal share between truck and rail as a function of freight and shipment characteristics. However, none of the previous work predicts parcel shipment split in the e-commerce environment.

Traditional optimization methods for decision-making suffer from the curse of dimensionality. Many literature focus on the operations research modeling aspects of reducing dimensionality, however, none of the previous work integrates data analytics into optimization process to either reduce dimensionality or speed up optimization run time. Recently, data driven optimization has become an emerging topic by combining data analytics with optimization. \cite{Calafiore:2006}  have proposed data-driven methods for chance-constrained and distributionally robust problems. \cite{Bertsimas:2015} propose a novel schema for utilizing data to design uncertainty sets for robust optimization using statistical hypothesis tests. Their approach is flexible and widely applicable, and robust optimization problems built from it are computationally tractable, both theoretically and practically. However, this track of literature only focuses on using data analytics models to improve the robustness of stochastic optimization, none of them investigates the potential of using data analytics to reduce dimensionality.
\section{Methodology}

\begin{figure}
	\centering
	\subfigure{\label{fig:overview}\includegraphics[width = 0.45\textwidth]{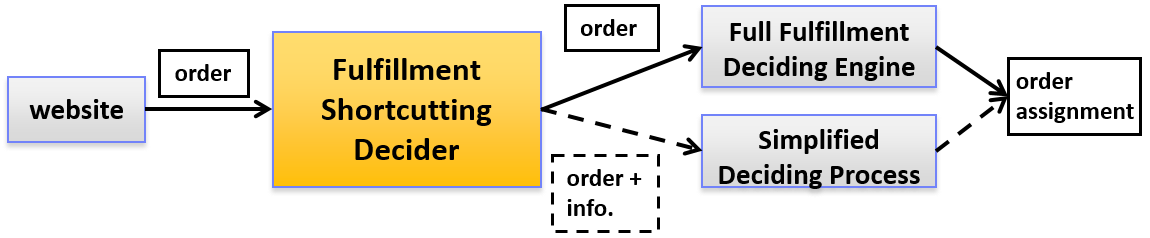}}
	\subfigure{\label{fig:splitcase}\includegraphics[width = 0.45\textwidth]{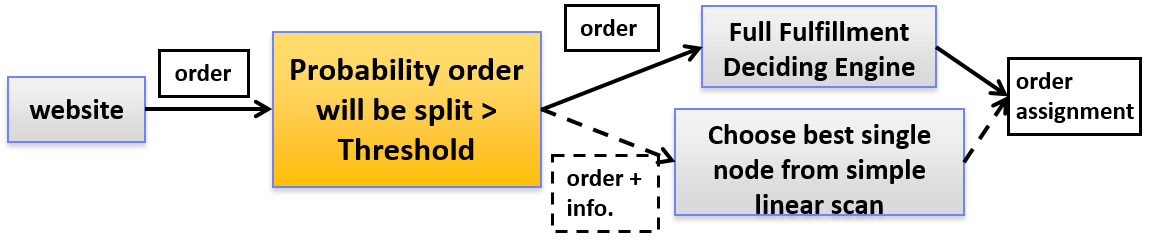}}
	\caption{Left: general short-cutting process - the dotted line shows the potential shortcut path taken, if the shortcutting decides short-cutting is possible. Right: order split deciding case.}
\end{figure}

Figure \ref{fig:overview} illustrates the key idea of our proposed general shortcutting framework. When an order comes in from the website, instead of going directly to the fulfillment deciding engine it is first passed to a new shortcutting decider component.  This component decides if the order corresponds to one of the cases for which a simpler deciding process can be used - with this decision system possibly continuously updated in real time and based on historical data.  For the split order deciding case, the decision checked is if the order would likely be split if it were to be decided by the regular fulfillment engine (Figure \ref{fig:splitcase}).  If the chance of split is above a specified probability threshold (which can also be configured based on the data) then the order is passed to the regular deciding engine, otherwise it is assumed the best decision would be to not split the order, so that then the decision can be determined by a simple linear scan of the nodes to find the best cost single node.  

In order to determine if an order is likely to be split or not, we model the problem as a machine learning / data mining one.  The general process for building the predictive model for determining order splitting is the following.  First, collect data from historical orders including whether or not the fulfillment engine decided to split the order, the general costs and characteristics of the order that are also available to the fulfillment engine, and optionally any other potentially relevant and readily available characteristics of the order such as node and item characteristics.   We say in the historical data, for the $i$th order out of $n$ orders, the ``label'' $y_i$ for that order was whether or not the order was split, taking value $1$ if the order was split and $0$ otherwise.  

Next, generate features from each order that hopefully capture the key characteristics of the order in a common representation (feature) space.  I.e., we cannot easily directly use the costs and characteristics associated with an order for modeling as each order has a different number of items and a different number of candidate nodes for items.  Instead we derive features that generalize across different orders like minimum shipping cost across items in the order, or average inventory level across candidate nodes.  We denote the $j$th feature of order $i$ as $x_{ij}$ so that if we derive $p$ features $\vec{x_i}$ is a $p$-dimensional vector. Then the task becomes to estimate a predictive model, function $f(\vec{x})$, mapping a feature vector $\vec{x}$ to a probability for the value of the label $y$, from the historical data.   Once this model is fit to the data, we can also analyze the impact of different thresholds on performance to arrive at an acceptable trade-off between the number of cases we can shortcut and the accuracy of our shortcutting decision (as illustrated with the precision-recall results in section \ref{sec:experiment}).  Then this predictive model, potentially  simplified and reduced to using features determined to be of key importance from the modeling, is used in the live system: an order arrives, its feature vector in the common representation space is generated and passed through the model, and a probability of splitting is output and checked against the threshold.  Additionally this model can be updated in an online fashion as orders are decided, factor in recent order history and decisions, and be tailored for different seasons, e.g., by using multiple models.

In order to fit the probabilistic mapping, $f$, to the historical data, we can use any number of machine learning / data mining methods that provide probabilistic outputs.  In general these methods try to minimize some loss function computed over the empirical data in combination with a trade-off regularization component intended to reduce the variance of the model estimation and prevent over-fitting.  E.g.: 
$
\label{eq:gloss}
\begin{array}{l ll}
\displaystyle\argmin_{f} & \sum_{i=1}^{n}L(f(\vec x_i), y_i) + R(\lambda, f)
\end{array}
$ 
where $L$ is a loss function penalizing differences in predicted labels from actual labels, such as logistic loss as described in section \ref{sec:experiment}, $R$ is some regularization function, such as an $L1$-norm penalty on the weight parameters of a regression model, and $\lambda$ is a hyper-parameter that trades off between the two components and is set from the data via model selection like cross-validation. In this work we use three standard modeling methods in our experiments, lasso ($L1$-norm regularized) logistic regression, decision trees, and boosted decision tree stumps - LogitBoost.  For reference on these methods and model estimation and selection refer to \cite{hastie2009elements}.
\section{Experimental Study} \label{sec:experiment}

We studied the viability of the shortcutting approach through evaluating split order prediction on a real world set of order data.  The following sections describe the data set used, the experiment protocol, and the results of the experiment.  

\textbf{Data Set:}  
In 2014, IBM Research developed a pilot solution for online order fulfillment optimization for a large US-based retailer with 1000+ stores and a few fulfillment centers. The retailer, like many others, is using "ship from store" (SFS) to fulfill online orders directly from its stores along with its fulfillment centers. SFS has many benefits such as proximity to customers, relieving load on fulfillment centers especially during the peak season, and using slow moving store inventory more effectively via online order fulfillment. The retailer in our case divided items into eligible for SFS and non-eligible for SFS categories. SFS-eligible items can be shipped from stores as well as fulfillment centers, whereas SFS-non-eligible items can only be shipped from fulfilment centers. The pilot solution by IBM Research used multi-objective optimization to simultaneously optimize multiple conflicting business objectives and find an optimal assignment for an online order. An assignment includes, for each ordered item, a choice of node(s) and quantities. 

In this paper we used this pilot solution to generate optimal assignments (and intermediate data extracts for feature learning) for 98,801 online orders during the Thanksgiving-Christmas season in 2013. On average each order in this sample consisted of 3.1 items, out which 1.8 are SFS-eligible and 1.3 are SFS-non-eligible. The optimizer pilot solution in this case was used with two conflicting business objectives - minimizing shipping cost and maximizing potential savings by using items in stores which are in less demand and will eventually go on clearance. We used the pilot solution to extract data for each order at various levels: order level, item level, node level and item-node level. At the order level, the information extract included items, order quantity for each item and order destination. At the item level we extracted information such as item shipping weight, price, cost, eligibility for clearance due to low demand, etc. At the node level, we extracted information on node type (store or fulfillment center) and shipping distance to the destination. At the item-node level, the extract included available inventory, shipping cost assuming single node assignment, sales so far and potential saving associated with clearance if any.  From these order characteristics, we generated 70 different efficiently-computable (at most linear in the number of candidate nodes per item) features we thought could be indicative in some way of order splitting.  For example, general features like total order quantity and weight, and item and node specific features like the maximum estimated clearance-related savings possible.

Of the 98,801 orders, 73,923 orders (74.82 \%) orders were not split.  There were 29,736 single item orders (30.1\% of all orders) which were removed from our analysis - since no split is possible for those orders we can immediately shortcut such orders without any prediction.  There were then 69,065 multi-item orders, equal to 69.9\% of all orders.  Among these, 44,187 were not split, equal to 63.98 \% of multi-item orders.

\textbf{Experiment Protocol:} 
\begin{wrapfigure}{r}{0.8\linewidth}
	\centering
	\subfigure{\label{fig:quant}\includegraphics[width = 0.25\textwidth]{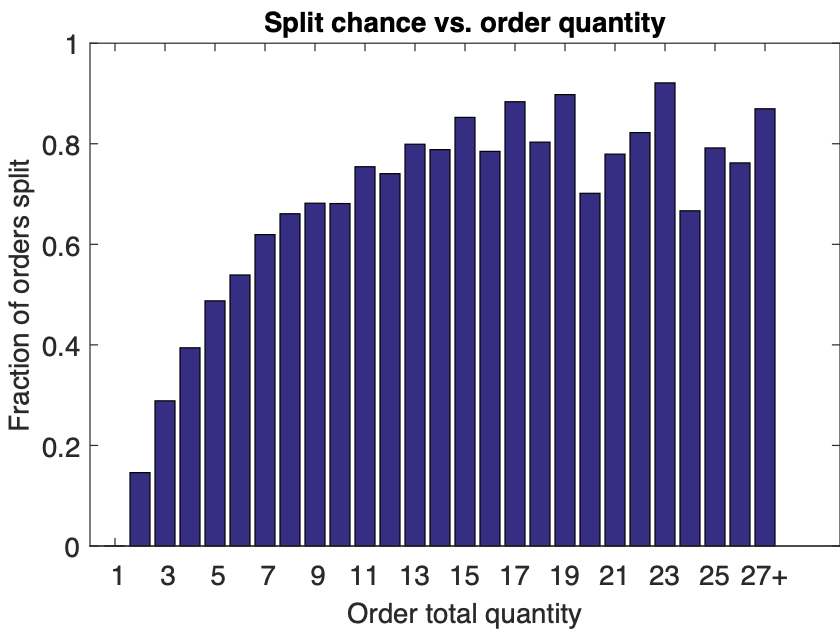}}
	\subfigure{\includegraphics[width = 0.25\textwidth]{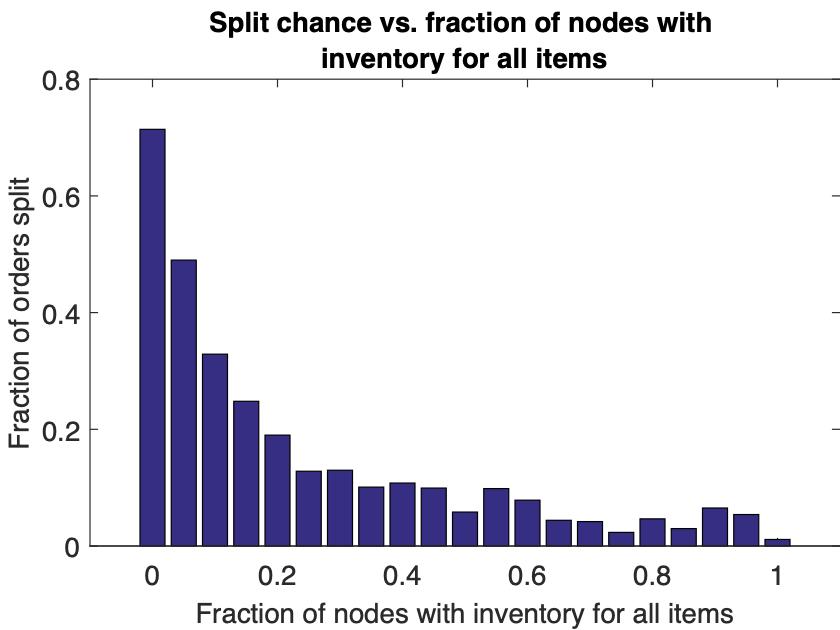}}
	\subfigure{\label{fig:sfs}\includegraphics[width = 0.25\textwidth]{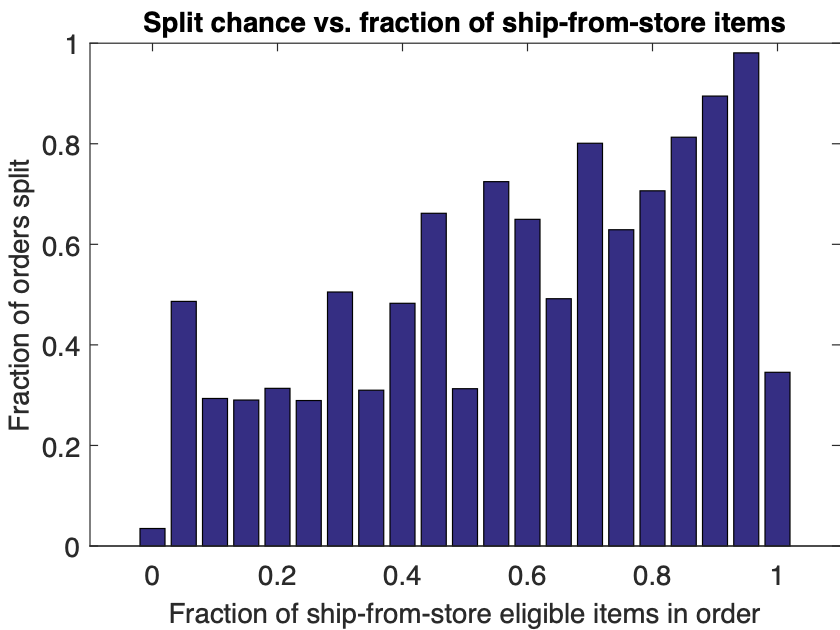}}
	\caption{Histograms capturing the distribution of order split chance vs. feature values, to give an idea about features' relationships to the label (order splitting). Split chance vs., from left to right, total quantity of order, fraction of candidate nodes with inventory for all items, and fraction of the order that is eligible for shipping from store.}
	\label{fig:featurecorr}
\end{wrapfigure}
To evaluate the performance of each predictive modeling method, we used 5 repetitions of 10-fold cross-validation \cite{hastie2009elements} on the complete set of orders.  For each training set in a fold, we use another application of 10-fold cross validation to select the regularization parameters for each method.  Note that this essentially assumes that the data sample is representative of the same data we would apply the prediction to.  We feel this is a relatively fair representation in this case, since the features of the order are very generic and we can think of the application as being for the case of applying the approach to other orders in the same day or similar peak days. Applying and potentially adapting the models to other scenarios (days, seasons) - which may possibly correspond to different underlying data distributions - is an area of future work.

We report both average accuracy and log loss as overall measures of performance on the complete set of orders in each test set - log loss to incorporate the confidence of the predictions.  Specifically, given each label $y_i$, taking values 1 if order $i$ was split, 0 if not split, and a model's corresponding prediction probability of split $p_i$ then accuracy is given by $\sum_i I(y_i = I(p_i > 0.5))$ and log loss by $-\sum_i y_i\log p_i + (1-y_i)\log(1-p_i)$.  Additionally we recorded accuracy at varying confidence levels and use this to also generate precision-recall curves for each method.

\textbf{Experiment Results:}  
The result of applying order split prediction to the peak day orders and assignments are given in the following sub sections.  First, we provide some preliminary analysis of the characteristics of the data  that give credence to the idea that some sort of prediction is possible - showing some examples of clear trends between key features and the split outcome.  Then we report the overall performance of the predictive approach as well as the trade-off between precision and recall to demonstrate the viability of using the approach in a real system.  Finally, we break down the learned predictive models to give some insight into the problem and solutions found.

\textbf{Data Analysis:} 
Figure \ref{fig:featurecorr} shows some examples of correlations between some presumed key features and order split chance.  Each feature is shown on the x-axis and its values in the data broken into bins, then on the y-axis the fraction of split orders for all orders falling in that bin is given.  This gives an idea of the correlation or trend between features and the order split outcome.  There are some clear trends here, for example, the chance an order is split increases with increase in the total quantity of items in the order (figure \ref{fig:quant}), although the relationship does not appear to be completely smooth or linear.  In other cases, e.g., figure \ref{fig:sfs}, there seems to be some correlation but the relationship is less clear.

\textbf{Prediction Performance:}
\begin{wraptable}{r}{0.70\textwidth}
	\centering
	\caption{Mean $\pm$ std. dev. of accuracy and log loss with best scores shown in bold. \label{tab:accres}}
	\begin{scriptsize}
		\begin{tabular}{|l|l|l|l|l|}
			\hline
			&\textbf{L1 logistic regression}&\textbf{Decision tree}&\textbf{LogitBoost}&\textbf{Ensemble}\\\hline
			\textbf{log loss}&0.2120$\pm$0.0056&0.1731$\pm$0.0079&0.1653$\pm$0.0060&\textbf{0.1574$\pm$0.0063}\\\hline
			\textbf{accuracy}&0.9244$\pm$0.0033&0.9409$\pm$0.0034&0.9407$\pm$0.0032&\textbf{0.9443$\pm$0.0036}\\\hline
		\end{tabular}
	\end{scriptsize}
\end{wraptable}

\begin{wrapfigure}{r}{0.5\linewidth}
	\centering
	\includegraphics[width = 0.45\textwidth]{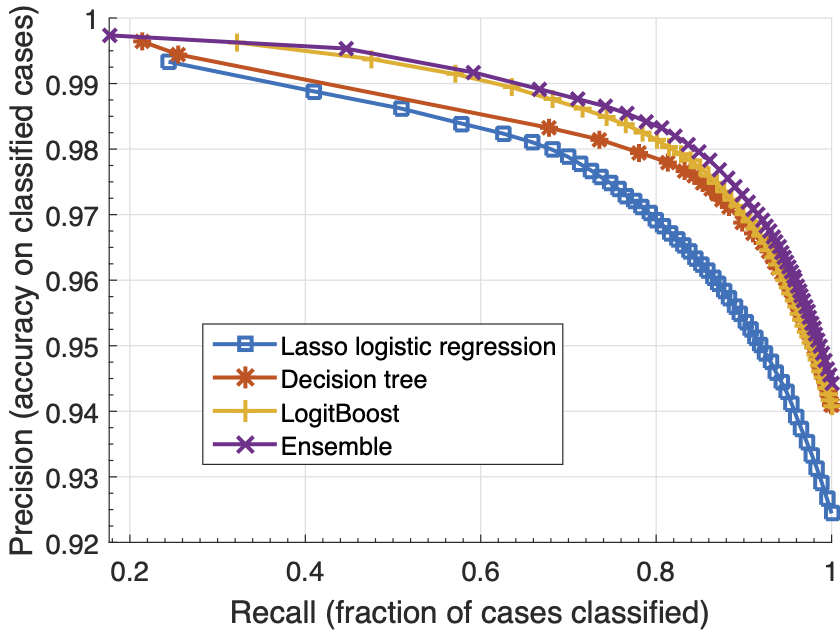}
	\caption{Precision vs. recall from varying confidence threshold.}
	\label{fig:prcurve}
\end{wrapfigure}

Overall performance measures are given in Table \ref{tab:accres}, and precision-recall curves, showing the accuracy vs the fraction of test cases classified for varying confidence threshold are given in Figure \ref{fig:prcurve}.  We found we were able to achieve high accuracy at predicting order splits - which suggests the proposed fulfillment short-cutting approach has real promise and that the proposed order split decider could be a key component.  The ensemble approach, created by simply averaging the prediction probabilities from the decision tree and boosting methods, achieved the best performance overall - $0.157$ log loss, $94.4\%$ accuracy and $99.17\%$\ accuracy when classifying test cases it was at least $97\%$ confident on, equal to $59.23\%$ of test cases.  Also the decision tree approach by itself performed nearly as well - which is an extra plus since decision trees can also provide interpretable rules that can be easily implemented in existing systems. 

\textbf{Learned Model Analysis:}
We also looked at the relative importance of the different features across methods and analyzed the learned models. Overall the top 3 features across methods were, with respect to the items of an order and its candidate fulfillment nodes: estimated max. possible clearance-related savings, estimated max. per-item clearance-related savings restricted to nodes with inventory for all items, and the fraction of nodes with inventory for all items, with shipping cost and total weight features also in the top 10.  All the methods agreed on the top predictor and generally had large overlap in key features.  

For decision trees, across all cross-validation iterations, the average tree depth was 32.9, the average number of cases per leaf was 35.4, and the average number of nodes was 3,511.3.  We examined the rules learned in the decision trees.  Due to space constraint, we just give one example decision sequence for one of the shorter sequences.  This sequence seems to address the case of a combination of a very large number of candidate nodes with only a small fraction having inventory for all items: "if the estimated max. possible clearance-related savings is greater than or equal to 4.6886, and if the fraction of nodes with inventory for all items is less than 0.00247219, and if the estimated best no-split total cost per item is less than < 5.72585, and if the number of candidate nodes is greater than 410, then the order should not be split."   

\section{Conclusion and Future Work}
We proposed a new approach to ease the burden on fulfillment deciding engines based on introducing a novel short-cutting component in the deciding process. We demonstrated the viability of a particular use-case of this approach, via order split prediction, which can allow simple fulfillment deciding for cases where an order is not split.  In the future we plan to explore other motivating short-cutting use cases, as mentioned in the introduction, as well as deriving simple, easily interpretable rules for short-cutting.  Additionally we plan to investigate the broader applicability of the approach and address potential issues that may arise such as covariate shift.  Finally, we plan to also incorporate other factors in the short-cutting prediction tasks, in particular other costs, and time and auto-regressive factors, and investigate and evaluate the short-cutting approaches from a total cost improvement perspective.

\setlength{\bibsep}{1pt} 
\renewcommand{\bibfont}{\scriptsize} 
\bibliographystyle{abbrvnat}
\bibliography{INFORMS}

\end{document}